\documentclass[letterpaper]{article} 
\usepackage{aaai24}  
\usepackage{times}  
\usepackage{helvet}  
\usepackage{courier}  
\usepackage[hyphens]{url}  
\usepackage{graphicx} 
\usepackage{natbib}  
\usepackage{caption} 
\usepackage{amsmath}
\usepackage{amssymb}
\usepackage{pifont}
\newcommand{\cmark}{\ding{51}}%

\usepackage{textcomp}

\usepackage{booktabs}
\usepackage{multirow}
\usepackage{makecell}
\usepackage{romannum}

\usepackage{algorithm}
\usepackage{algorithmic}

\usepackage{overpic} 
\usepackage{contour}
\usepackage{color}

\usepackage{float} 
\usepackage{subcaption}

\newcommand{\myPara}[1]{\vspace{2pt}\noindent\textbf{#1}}
\newcommand{\nameofmethod}{Polyper}
\newcommand{\nameofrefine}{boundary sensitive attention}
\newcommand{\nameofedge}{region separation}

\usepackage{newfloat}
\usepackage{listings}
\DeclareCaptionStyle{ruled}{labelfont=normalfont,labelsep=colon,strut=off} 
\floatstyle{ruled}
\newfloat{listing}{tb}{lst}{}
\floatname{listing}{Listing}
\title{Polyper: Boundary Sensitive Polyp Segmentation}
\author{
    Hao Shao\textsuperscript{\rm 1} \quad\quad
    Yang Zhang\textsuperscript{\rm 2}  \quad\quad
    Qibin Hou\textsuperscript{\rm 1}\thanks{Corresponding author.}
}
\affiliations{
    \textsuperscript{\rm 1} VCIP, School of Computer Science, Nankai University\\
    \textsuperscript{\rm 2} Department of Genetics and Cell Biology, College of Life Sciences, Nankai University\\
}

\usepackage{bibentry}

\begin{document}

\maketitle

\begin{abstract}

We present a new boundary sensitive framework for polyp segmentation, called~\nameofmethod{}.
%
%
Our method is motivated by a clinical approach that seasoned medical practitioners often leverage the inherent features of interior polyp regions to tackle blurred boundaries.
Inspired by this, we propose explicitly leveraging polyp regions to bolster the model's boundary discrimination capability while minimizing computation. 
Our approach first extracts boundary and polyp regions from the initial segmentation map through morphological operators.
Then, we design the boundary sensitive attention that concentrates on augmenting the features near the boundary regions using the interior polyp regions's characteristics to generate good segmentation results.
Our proposed method can be seamlessly integrated with classical encoder networks, like ResNet-50, MiT-B1, and Swin Transformer.
To evaluate the effectiveness of~\nameofmethod{}, we conduct experiments on five publicly available challenging datasets, and receive state-of-the-art performance on all of them.
Code is available at \it{\url{https://github.com/haoshao-nku/medical_seg.git}}.
\end{abstract}


\section{Introduction} \label{sec:intro}
Colon polyps are protruding growths within the colon mucosa, exhibiting considerable variability in shape, texture, and color~\cite{pooler2023growth}.
Importantly, colon polyps are recognized as precancerous lesions closely associated with the development of colon cancer~\cite{djinbachian2020rates}. 
Consequently, there is a pressing need to enhance both the efficiency of early detection and the accuracy of polyp contour segmentation.
Polyps pose diagnostic challenges during colonoscopy due to the inconspicuous borders and low contrast. 
In their initial stages, polyps often manifest smaller dimensions, resulting in less defined margins that exacerbate detection difficulties.

\begin{figure}[htbp]
    \centering
    
    \begin{minipage}{0.49\linewidth}
        \centering
        \begin{overpic}[width=1.05\linewidth]{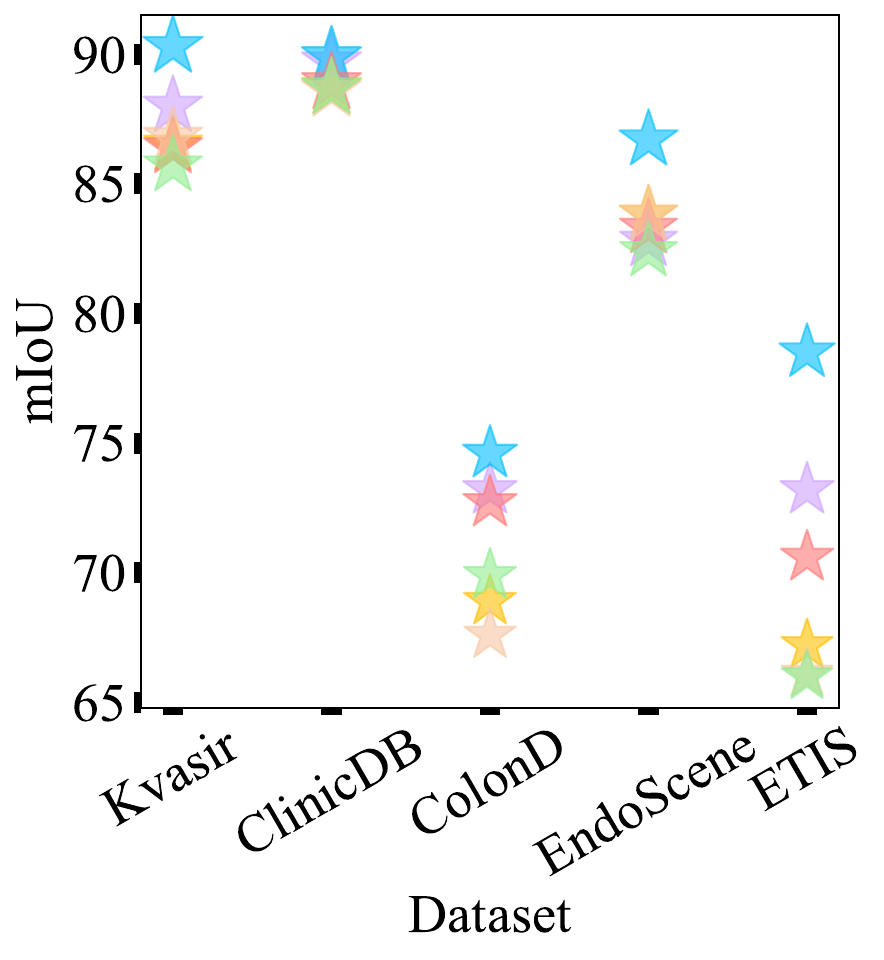}
        \end{overpic}
    \end{minipage}
    \begin{minipage}{0.49\linewidth}
        \centering
        \begin{overpic}[width=1.05\linewidth]{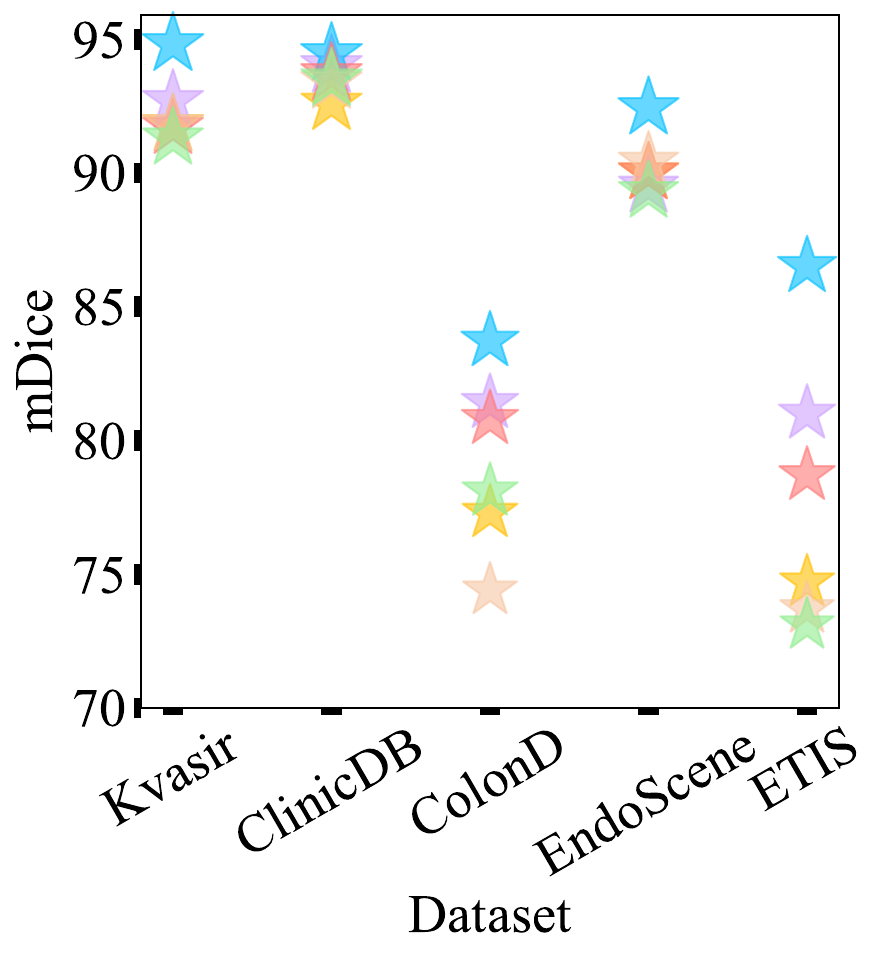}
        \end{overpic}
    \end{minipage}
    
    \begin{minipage}{0.49\linewidth}
        \centering
        \begin{overpic}[width=1\linewidth]{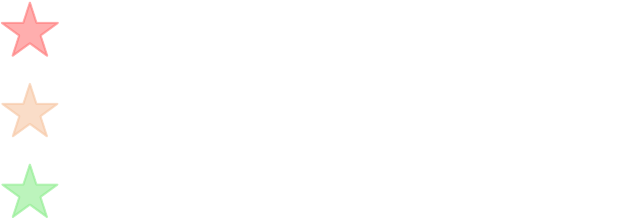}
    \put(15, 1.3){{\fontsize{8}{10}\selectfont \textcolor{black}{Polyp-PVT~\cite{dong2021polyp}}}}
    
    \put(15, 14){{\fontsize{8}{10}\selectfont \textcolor{black}{TransFuse (Zhang et al. 2021)}}}
    
    \put(15, 27){{\fontsize{8}{10}\selectfont \textcolor{black}{TransUNet~\cite{lin2022ds}}}}
    
        \end{overpic}
    \end{minipage}
    \begin{minipage}{0.49\linewidth}
        \centering
        \begin{overpic}[width=0.85\linewidth]{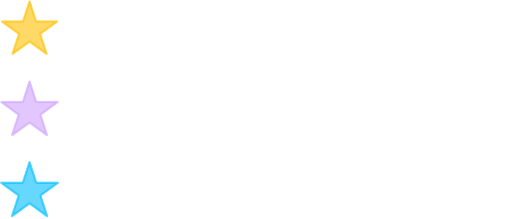}
    \put(15, 32){{\fontsize{8}{10}\selectfont \textcolor{black}{CaraNet~\cite{lou2022caranet}}}}
    \put(15, 16){{\fontsize{8}{10}\selectfont \textcolor{black}{SegT (Chen et al. 2023)}}} 
    \put(15, 1){{\fontsize{8}{10}\selectfont \textcolor{black}{\nameofmethod{} (Ours)}}} 
        \end{overpic}
    \end{minipage}
    
    \caption{Comparison to state-of-the-art methods on five popular datasets.}
    \label{fig:intro}
\end{figure}

To address this challenge, one of the current research trends is to maximize the integration of features at various scales to preserve as many boundary details as possible.
Typically, Wu et al.~\cite{wu2021precise} introduced semantic calibration and refinement techniques to bridge the semantic gap between feature maps at different levels, resulting in good polyp segmentation maps.
Similarly, SwinE-Net~\cite{park2022swine} refines the multi-level features extracted from both CNN and Swin Transformer architectures using multi-dilation convolutions and multi-feature aggregation blocks.
While this approach can deal with well-margined polyps in the mid to late stages of a well-margined lesion, it struggles to effectively handle early polyps with lower edge contrast.

Another popular strategy is to generate a polyp mask to coarsely localize the polyp and then improve the quality of the segmentation maps by enhancing the semantic features around potential boundaries.
As an early attempt, SegT~\cite{chen2023segt}  highlights the boundaries of the polyp area by assessing the disparity between the foreground and background regions. 
However, accurately capturing the polyp boundaries poses a challenge due to the ambiguity caused by blending polyp boundaries with the surrounding mucosa. 
Relying solely on the synergistic effect between foreground and background information may not lead to accurate polyp segmentation.
CaraNet \cite{lou2022caranet} leverages the long-distance interaction of axial attention to calculate the pairwise affinity from a global perspective.
The roughly estimated polyp regions are removed from the deep features, and then features at different scales are utilized to supplement the boundary details to produce good segmentation results.
However, axial attention may not consistently benefit the network because it risks inadvertently excluding crucial boundary information~\cite{thanh2022colonformer}.
Hence, the methods above cannot sufficiently address the issue of blurred boundaries.

In endoscopic screening, skilled medical practitioners often utilize the polyp characteristics of non-boundary regions to address the challenge of boundary blurring.
Drawing inspiration from this observation, our method first generates an initial segmentation map and employs morphological operators to partition the  polyp into boundary and non-boundary regions.
We then leverage the semantic features extracted from the non-boundary regions to refine the boundary regions with a novel boundary sensitive attention module, which can take advantage of both global and local features to identify the real polyp boundaries.
%
%
%

Our method, called~\nameofmethod{}, is simple, easy to follow, and suitable for practical medical scenarios.
%
%
%
We conduct a series of experiments on five widely used datasets to evaluate ~\nameofmethod{}.
%
As shown in Fig.~\ref{fig:intro},~\nameofmethod{} outperforms previous methods in all datasets in terms of mDice and mIoU scores.
Typically, because of the heterogeneity of polyp features, previous works mostly do not perform well
for small polyps in the early growth stage.
%
In the subsequent ablation study, we verify that our method performs well on small polyps.
%
%

Our main contributions can be summarized as follows:
\begin{itemize}
    \item We present a novel module, named boundary sensitive attention, which can model the relationships between the boundary regions and interior regions of polyps to augment the features near the boundary regions by capitalizing on the inherent characteristics of the interior regions.
    \item We design a novel decoder for polyp segmentation composed of two distinct stages: potential boundary extraction and boundary sensitive refinement. This decoder helps us effectively identify the real polyp boundaries, resolving the challenge of boundary blurring in endoscopy.
    \item We evaluate the proposed~\nameofmethod{} on five popular polyp segmentation datasets and set new records on almost all the benchmarks.
\end{itemize}

\section{Related Work}

\myPara{Architectures for Medical Image Segmentation.}
CNN is one of the most widely used deep neural network architectures in medical image segmentation. 
A typical example should be U-Net~\cite{ronneberger2015u}, one of the most classic networks.
Attention U-Net~\cite{oktay2018attention} introduces a novel attention gate mechanism that empowers the model to selectively focus on targets of diverse shapes and sizes.
Res-UNet~\cite{xiao2018weighted} incorporates a weighted attention mechanism to improve segmentation performance.
R2U-Net~\cite{alom2018recurrent} ingeniously merges the strengths of residual networks~\cite{he2016deep} and U-Net.
KiU-Net~\cite{valanarasu2020kiu} proposes an innovative structure that leverages under-complete and super-complete features to enhance the segmentation of lesion regions with small anatomical structures.
AttResDU-Net~\cite{khan2023attresdu} incorporates attention gates on the skip connections and residual connections in the convolutional blocks.
Recently, there has been a surge of interests in utilizing Transformers~\cite{huang2022rtnet} for polyp segmentation.
SwinMM~\cite{wang2023swinmm} develops a cross-viewpoint decoder that aggregates multi-viewpoint information through cross-attention blocks.
MPU-Net~\cite{yu20233d} aims to achieve precise localization by combining image serialization with a positional attention module, enabling the model to comprehend deeper contextual dependencies effectively.
Segtran~\cite{li2021medical} presents a novel  Squeeze-and-Expansion transformer.


\begin{figure*}
  \centering
  \setlength{\abovecaptionskip}{2pt}
 \begin{overpic}[width=0.9\linewidth]{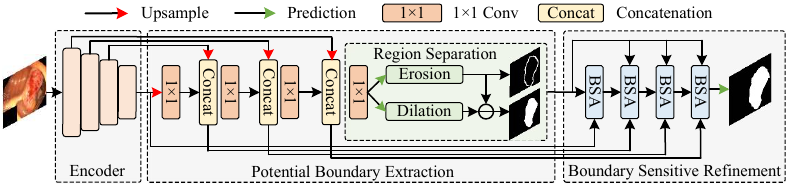}
  \end{overpic}
  \caption{Overall architecture of \nameofmethod{}.
  We use the Swin-T from Swin Transformer as the encoder.
  The decoder is divided into two main stages.
  The first potential boundary extraction (PBE) stage aims to capture multi-scale features from the encoder, which are then aggregated to generate the initial segmentation results.
  Next, we extract the predicted polyps' potential boundary and interior regions using morphology operators.
  In the second boundary sensitive refinement (BSR) stage, we model the relationships between the potential boundary and interior regions to generate better segmentation results.}
  \label{fig:backbone}
\end{figure*}

\myPara{Refinement Methods.}
One avenue for refinement is to maximize the utilization of features at different scales.
Wu et al.~\cite{wu2021precise} employed semantic calibration and refinement techniques to bridge the semantic gap between different levels of feature mapping.
SwinE-Net~\cite{park2022swine} refines and enhances the multi-level features extracted from CNN and Swin Transformer through multi-dilation convolutions and multi-feature aggregation blocks.
FTMF-Net~\cite{liu2023ftmf} presents a Fourier transform multiscale feature fusion network for segmenting small polyp objects.
Another kind of approaches involves targeting specific areas for refinement.
PraNet~\cite{fan2020pranet} and CaraNet~\cite{lou2022caranet} both integrate the Reverse Attention module~\cite{chen2018reverse}, a specialized component that accentuates the boundaries between polyps and their surroundings.
Xie et al.~\cite{xie2020segmenting} introduced assisted boundary supervision as a guiding mechanism for refining glass segmentation, aiding in predicting uncertain regions around the boundaries.
In contrast to direct boundary feature enhancement, He et al.~\cite{he2021enhanced} advocated supervising the non-edge portion through a residual approach to attain finer edges.
Zhang et al.~\cite{zhang2020adaptive} proposed the Local Context Attention module to pass local context features from the encoder layer to the decoder layer, enhancing the focus on hard regions.
RFENet~\cite{fan2023rfenet} introduces a structure focus refinement module to facilitate fine-grained feature refinement of fuzzy points around the boundaries.
EAMNet~\cite{sun2023edge} considers edge detection and camouflaged object segmentation as an interlinked cross-refinement process.


\section{Method}

Accurately recognizing polyp boundaries from the surrounding mucosa is challenging due to the low contrast between the polyp and surrounding tissues. 
To address this issue, a prevalent strategy is to enhance the quality of the segmentation map by refining the semantic features near the potential boundaries.
However, the ambiguous nature of polyp boundaries mixed with the surrounding tissues often hinders accurate prediction of the polyp boundary regions.
In the context of endoscopic screening, seasoned medical practitioners often leverage the inherent features of polyps within non-boundary regions to effectively tackle the issue of blurred boundaries.
Motivated by this clinical approach, we present a new boundary sensitive framework, called~\nameofmethod{}.

Fig.~\ref{fig:backbone} provides an overview of~\nameofmethod{}. 
Like most previous works for polyp segmentation, we employ the classical encoder-decoder architecture.
The encoder plays a crucial role in extracting features at different scales and levels, enabling the model to capture coarse and fine details. 
We utilize the Swin-T from Swin Transformer~\cite{liu2021swin} as our encoder.
The decoder comprises two distinctive stages: potential boundary extraction and boundary sensitive refinement.
In the potential boundary extraction stage, the encoder's multi-scale features are aggregated to generate an initial prediction, which is used to extract the predicted polyps' potential boundaries and interior regions.
The boundary sensitive stage leverages the distinctive characteristics of the interior regions to enhance the model's precision by modeling the relationships between the potential boundary regions and the interior regions.
In what follows, we will describe these two stages in detail.

\subsection{Potential Boundary Extraction}

An overview of the potential boundary extraction stage is depicted in Fig.~\ref{fig:backbone}.
We predict the segmentation map using $1\times1$ convolution and employ morphology operators to extract the boundaries and interior polyp regions from the initial segmentation result.
This stage can be separated into feature aggregation and region separation.

%

\myPara{Feature Aggregation.} In the feature aggregation part, we use $1\times 1$ convolution and the concatenation operation to aggregate features of different scales. 
Given the features from the four stages of the encoder, denoted as $E_0, E_1, E_2$, and $E_3$\footnote{The resolutions are denoted as $ \frac{{H}}{4} \times \frac{{W}}{4} $, $\frac{{H}}{8} \times \frac{{W}}{8} $, $\frac{{H}}{16} \times \frac{{W}}{16} $, and $ \frac{{H}}{32} \times \frac{{W}}{32}$, respectively. ${H}$ and ${W}$ are the height and width of the input image, respectively.},
we first build a feature pyramid following~\cite{lin2017feature}.
%
Specifically, we resize feature maps $E_1, E_2$, and $E_3$ to ensure  they share the same size as $E_0$ through linear interpolation, yielding $E_1^\prime,E_2^\prime$, and $E_3^\prime$.
The formula for calculating feature map ${D_i}$ of the intermediate layer in each stage of the feature aggregation process is:
\begin{equation}
D_i=[\mathrm{Conv}_{1\times 1}(D_{i+1}), E_i^\prime],
\end{equation}
where $i \in \left \{ 0,1,2,3 \right \}$ and $[\cdots]$ means the concatenation operation.
Here, $D_3$ is equivalent to $E_3^\prime$ and $E_0^\prime$ is equivalent to $E_0$.
$\mathrm{Conv}_{1\times 1}$ means $1\times 1$ convolution.

\myPara{Region Separation.} 
We introduce a region separation module to separate the boundaries and the interior polyp regions from the initial segmentation map.
Specifically, given the output $D_0$ of the last stage of the feature aggregation, the initial segmentation mask $f_m$ is obtained by a $1\times 1$ convolution.
Then, we utilize the erosion operator ($E$) and the dilation operator ($D$)\footnote{Please refer to Chapter 9 of the Digital image processing~\cite{gonzales1987digital} for more details.} to separate the boundary and interior regions from the initial segmentation mask.
At each iteration, the mask edge erodes or expands by one pixel.
These regions offer  essential guidance for the subsequent refinement process.
The separation process can be written as:
\begin{align}
P_{CR} &= E(f_m)\times T, \\
P_{BR} &= (D(f_m)\times T- P_{CR}),
\end{align}
where $P_{CR}$ is the interior regions while $P_{BR}$ denotes the potential boundary regions.
%
Here, $T$ is the number of operations for operator $D$.
It is noteworthy that the total number of operations of operator $D$ and operator $E$ is the same.

\subsection{Boundary Sensitive Refinement}

As mentioned in the introduction section, the diverse characteristics of polyps in different growth stages, including their shape, texture, and color, lead to significant challenges to the robustness of polyp segmentation methods.
%
%
To address this, we present the boundary sensitive refinement stage to refine the feature of boundary regions based on $P_{CR}$ and $P_{BR}$.

The boundary sensitive refinement stage can be illustrated in the right part of Fig.~\ref{fig:backbone}, which includes boundary sensitive attention module and full-stage sensitivity strategy.
We first extract features from the boundary, interior polyp, and background regions, respectively.
Then, we leverage the cross-attention mechanism to model the relationships between the boundary region and the interior polyp regions as well as the relationships between the interior polyp regions and the background regions.
%
This enables simultaneous encoding of global and local features to improve the quality of segmentation results with accurate boundaries.
After the above process is done, the features corresponding to different regions will be restored to their initial positions.
The goal is to ensure the efficient use of hardware resources.
To realize the process above, we introduce a novel \nameofrefine{} module.
In addition, deep features excel at capturing and conveying semantic information, while low-level features are good at representing complex geometric details.
We introduce a full-stage sensitivity strategy that tactically harnesses the strengths of both deep and shallow features.

\begin{figure}
	\centering
 \setlength{\abovecaptionskip}{2pt}
 \begin{overpic}[width=1\linewidth]{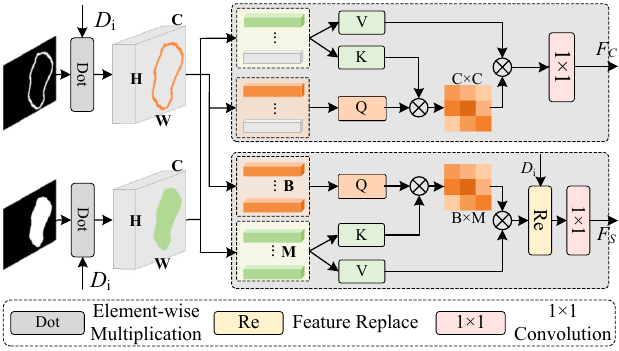}
  \end{overpic}
	\caption{Detailed structure of~\nameofrefine{} (BSA) module.
 This process is separated into two parallel branches, which systematically capitalize on the distinctive attributes of polyps at various growth stages, both in terms of spatial and channel characteristics.
 `B' and `M' indicate the number of pixels in the boundary and interior polyp regions within an input of size $H\times W$ and $C$ channels.
 }
	\label{fig:refine}
\end{figure}

\myPara{Boundary Sensitive Attention.} 
The structure of the \nameofrefine{} module is illustrated in Fig.~\ref{fig:refine}. 
It comprises two branches thoroughly exploring polyps' inherent characteristics by building spatial and channel attentions.
We first describe the working mechanism of the~\nameofrefine{} module on encoding spatial information.
%

Given the input $D_i$, the boundary region mask $P_{BR}$, and the interior polyp region mask $P_{CR}$,
we first perform element-wise product operations on $D_i$ over $P_{BR}$ and $P_{CR}$ to attain the features corresponding to the boundary region and the interior polyp region, denoted as $F_{BR}$ and $F_{CR}$, respectively.
%
To better discover the real polyp boundaries, we do not consider the background regions.
%
%
We treat $F_{BR}$ as the query matrix and $F_{CR}$ as the key and value matrices and compute the cross-attention between them as follows:
\begin{equation}
F_{S}=\mathrm{MHCA}_{S}(F_{BR}, {F_{CR}}, {F_{CR}}),
\end{equation}
where $\mathrm{MHCA}_{S}(\cdot, \cdot, \cdot)$ denotes multi-head cross-attention \cite{vaswani2017attention} along the spatial dimension.
%
%
This operation aims to more accurately mine regions that are real polyps by leveraging the priors of the interior polyp regions based on our observation mentioned at the beginning of this section.
In addition, as the cross-attention is computed over only $F_{BR}$ and $F_{CR}$ but not the background region, the computational cost is also low.
%
The results will then be put back to the corresponding positions of $D_i$.

We also consider using the background region to capture the variations and correlations between the background and boundary regions and between the background and internal polyp regions.
We use ${F_{BR}^\prime}$ and ${F_{CR}^\prime}$ to denote boundary region features and interior polyp region features that contain background information, respectively.
We treat ${F_{BR}^\prime}$ as the query matrix and ${F_{CR}^\prime}$ as the key and value matrices and compute the cross-attention between them as follows:
\begin{equation}
F_{C}=\mathrm{MHCA}_{C}(F_{BR}^\prime,F_{CR}^\prime,F_{CR}^\prime),
\end{equation}
where $\text{MHCA}_{C}(\cdot, \cdot, \cdot)$ denotes multi-head cross-attention along the channel dimension, following~\cite{yin2022camoformer,zamir2022restormer} to save computations.
The goal of this operation is to capture the consistency and correlation among the different regions from a global view. 
This helps better understand the overall structure of polyps.

The output of the proposed \nameofrefine{} module can be formulated as:
\begin{equation}
F_i=\mathrm{Conv}_{1\times 1}(F_{S})+\mathrm{Conv}_{1\times 1}(F_{C}) + D_i,
\end{equation}
where $\mathrm{Conv}_{1\times 1}$ is $1\times1$ convolution.

\myPara{Full-Stage Sensitive Strategy.} 
As depicted in the right part of Fig~\ref{fig:backbone}, the \nameofrefine{} module is utilized to effectively leverage the features at various scales for refining the boundary regions progressively.
The formulation of this full-stage sensitive strategy can be described as follows.
At the deepest stage, the refinement process is calculated as follows:
\begin{equation}
F_3=\mathrm{BSA}(D_3,P_{BR},P_{CR}),
\end{equation}
where $\mathrm{BSA}(\cdot, \cdot, \cdot)$ denotes the boundary sensitive attention module.
Furthermore, the refinement process in the subsequent stages can be defined as:
\begin{equation}
F_i=\mathrm{BSA}(\mathrm{Conv}_{1\times 1}(F_{i+1})+D_i,P_{BR},P_{CR}).
\end{equation}
Finally, we add a $1\times 1$ convolution to ${F_0}$ to generate the final predictions.
From the above descriptions, we can see that our decoder primarily contains $1\times1$ convolutions and simple cross-attention.
This makes our method efficient.

\section{Experiments}

\begin{table*}[tp]
  \centering
  \footnotesize
  \setlength{\abovecaptionskip}{2pt}
  \renewcommand\arraystretch{0.9}
  \setlength\tabcolsep{5.4pt}
  \caption{Comparisons with state-of-the-art methods.
 `Polyper w/o RS' means no region separation is used in potential boundary extraction, and the entire foreground region is refined instead of the boundary region.
  `Polyper w/o BSR' means not to refine the initial segmentation results.}
  \begin{tabular}{lccccccccccccccc}
  \toprule
  \multirow{2}{*}{Methods} &\multirow{2}{*}{Params (M)} &\multirow{2}{*}{Flops (G)} &\multicolumn{2}{c}{\makecell{Kvasir}} &\multicolumn{2}{l}{\makecell{ClinicDB}} &\multicolumn{2}{l}{\makecell{ColonDB}} &\multicolumn{2}{l}{\makecell{EndoScene}} &\multicolumn{2}{l}{\makecell{ETIS}}  \\
  \cmidrule(lr){4-5}\cmidrule(lr){6-7}\cmidrule(lr){8-9}\cmidrule(lr){10-11}\cmidrule(lr){12-13} 
  & & & \makecell{mDice}   & \makecell{mIoU}    & \makecell{mDice}   & \makecell{mIoU}     & \makecell{mDice}   & \makecell{mIoU}    & \makecell{mDice}   & \makecell{mIoU}     & \makecell{mDice}   & \makecell{mIoU}     \\ \midrule
U-Net  &24.56 &38.26 &81.80 &74.60 &82.30  & 75.50 &51.20 &44.40 &71.00 &62.60 &39.80 &33.50  \\
UNet++ &25.09 &84.30 &82.10 &74.30 &79.40 &72.90  &48.30 &41.00 &70.70 &62.40 &40.10 &34.40   \\
SFA &- &- &72.30 &61.10 &70.00 &60.70 &46.90 &34.70 &46.70 &32.90 &29.70 &21.70 \\
ACSNet &46.02 &29.45 &89.80 &83.80 &88.20 &82.60 &71.60 &64.90 &86.30 &78.70 &57.80 &50.90  \\
PraNet  &32.50 &221.90 &89.80 &84.00 &89.90 &84.90 &70.90 &64.00 &87.10 &79.70 &62.80 &56.70    \\
SANet &- &- &90.40 &86.40 &93.70 &88.90 &75.30 &67.00 &88.80 &81.50 &75.00 &65.40   \\
\midrule
TransFuse   &115.59  &38.73 & 91.80   & 86.80  & 93.40  & 88.60  & 74.40  & 67.60  & 90.40  & 83.80 & 73.70  & 66.10    \\
TransUNet   &105.28 &24.66 &91.30 &85.70 &93.50 &88.70 &78.10 &69.90 &89.30 &82.40 &73.10  &66.00 \\
HarDNet-MSEG  &33.80 &192.74 &91.20 &85.70 &93.20  &88.20 &73.10 &66.00 &88.70 &82.10 &67.70 &61.30   \\
DS-TransUNet  &177.44 &30.97  & 93.40  & 88.80  & 93.80  & 89.10  & 79.80     & 71.70   & 88.20 & 81.00 & 77.20 & 69.80    \\

SwinE-Net &- &- &92.00 &87.00 &93.80 &89.20 &80.40 &72.50 &90.60 &84.20 &75.80 &68.70   \\
Polyp-PVT  &- &-  &91.70 &86.40 &93.70 &88.90 &80.80 &72.70 &90.00 &83.30 &78.70 &70.60   \\
\midrule
CaraNet &46.64 &21.69 &91.80 &86.50 &93.60 &88.70 &77.30 &68.90 &90.30 &83.80 &74.70 &67.20\\
ColonFormer &52.94 &22.94 &92.40 &87.60 &93.20 &88.40 &81.10 &73.30 &90.60 &84.20 &80.10 &72.20   \\
SegT  &- &- &92.70 &88.00 &94.00 &89.70 &81.40 &73.20 &89.50 &82.80 &81.00 &73.20  \\
\midrule
\nameofmethod{} w/o BSR  &28.70 &37.42 &91.69 &85.16  &89.29 &81.86 &74.93 &66.00 &86.12 &80.25  &75.36 &66.77  \\
\nameofmethod{} w/o RS &29.82 &48.26  &91.97 &85.58 &88.35 &83.68 &75.60 &66.84 &87.69 &82.12  &82.36 &73.13\\
\nameofmethod{}  &29.11 &43.54 &\textbf{94.82} &\textbf{90.36} &\textbf{94.45} &\textbf{89.85} &\textbf{83.72} &\textbf{74.55} &\textbf{92.43} &\textbf{86.72} &\textbf{86.51} &\textbf{78.51}  \\ 
\bottomrule 
\end{tabular}
\label{POLYP}
\end{table*}

\begin{figure*}[t]
  \setlength{\tabcolsep}{0mm}
  \small
  \setlength{\abovecaptionskip}{2pt}
  \centering
  \begin{tabular}{ccccc}
    \includegraphics[width=0.2\textwidth]{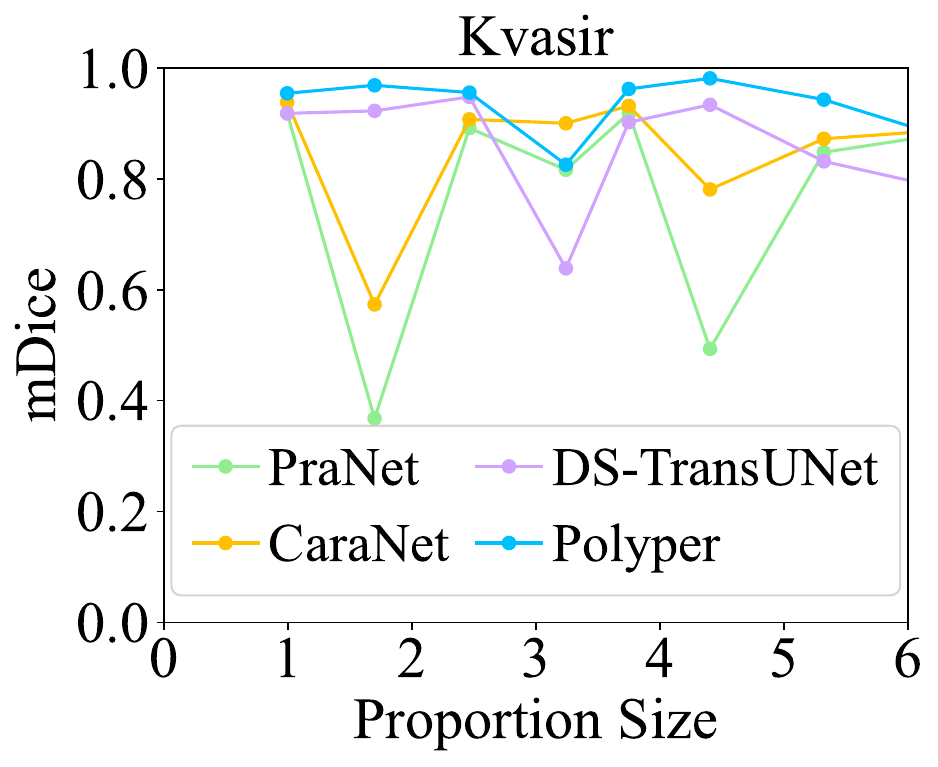}  
    & 
    \includegraphics[width=0.2\textwidth]{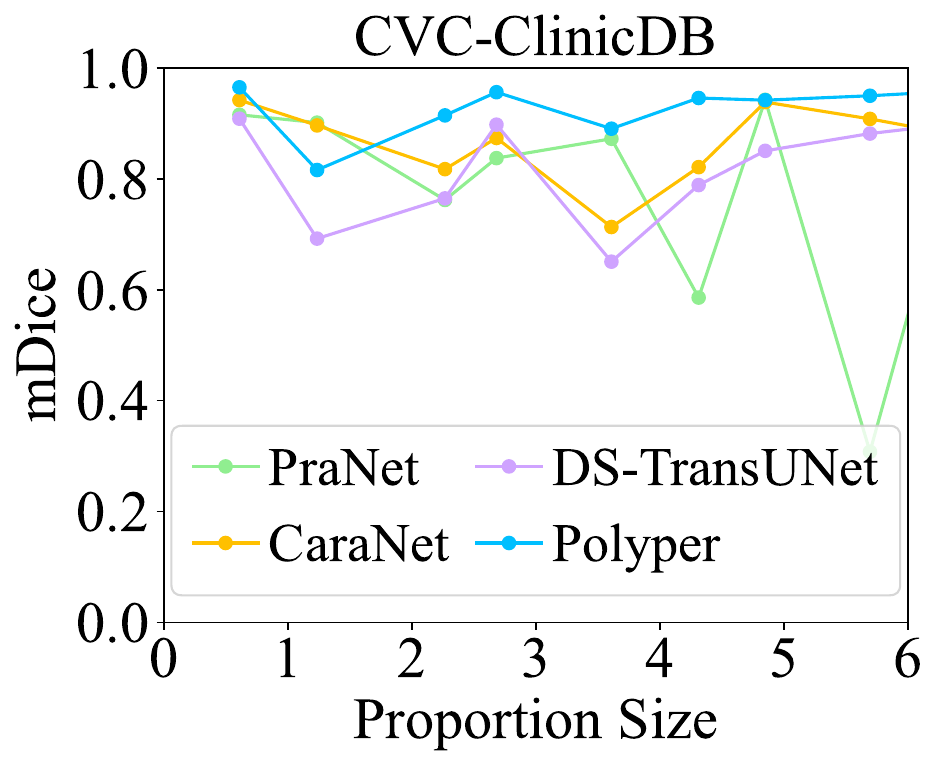}
    & 
    \includegraphics[width=0.2\textwidth]{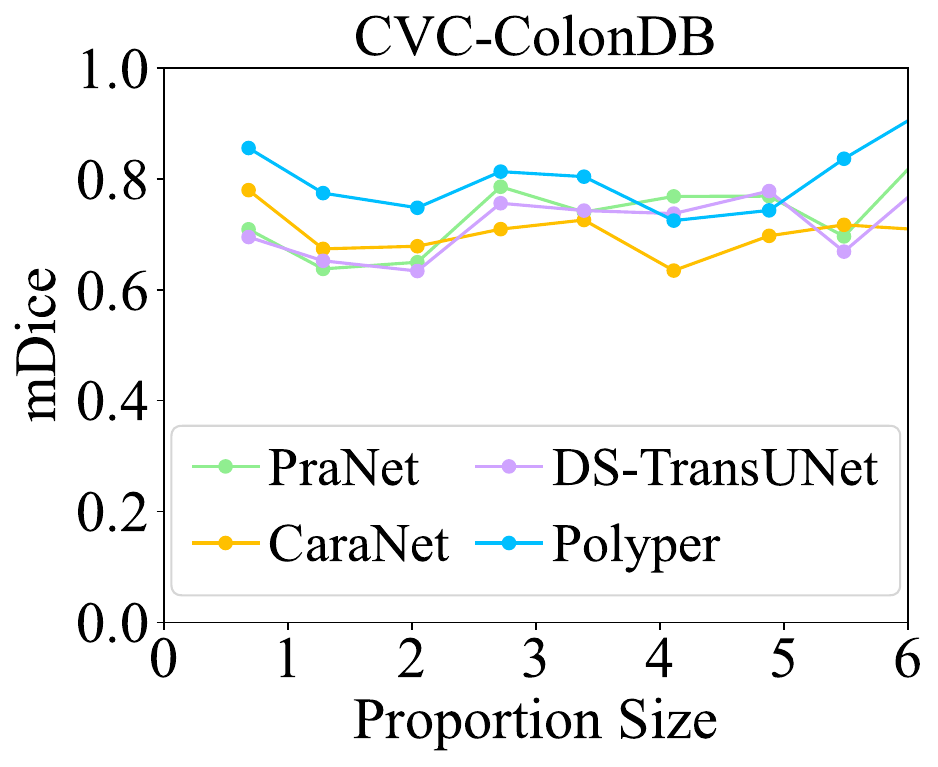}
    & 
    \includegraphics[width=0.2\textwidth]{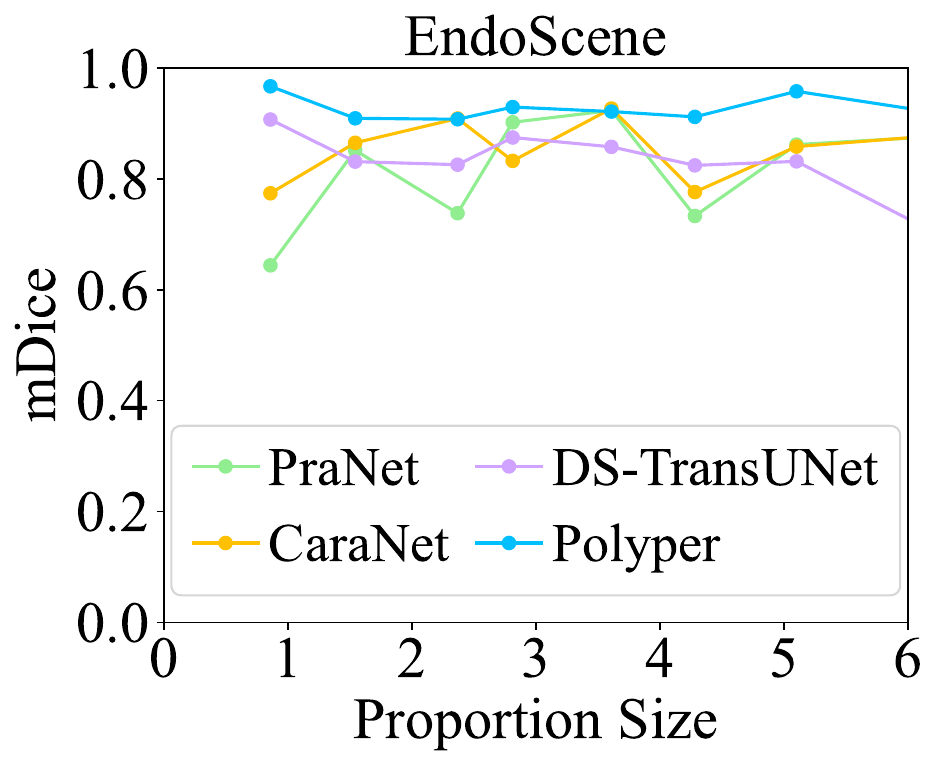}
    & 
    \includegraphics[width=0.2\textwidth]{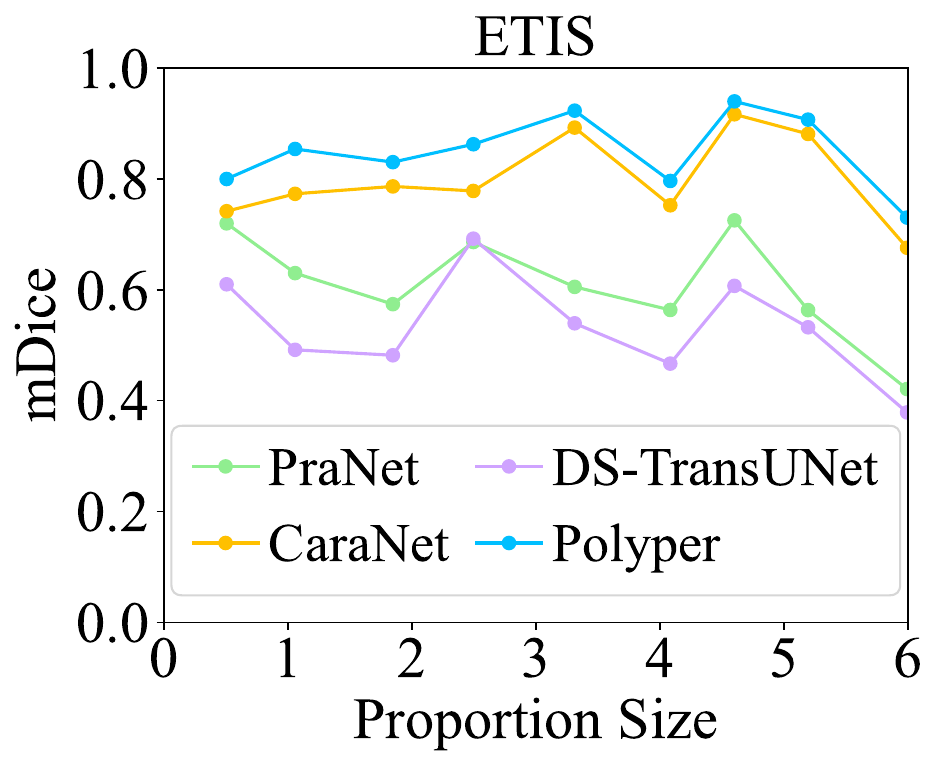}
    \end{tabular}
    
 \caption{Performance for small polyps on five dataset.
 `Proportion Size' is the ratio of the polyp's size to the entire image.
 }
\label{fig:smallpolyp}
\end{figure*}

\subsection{Dataset}
Following most previous works, the results are reported on the same datasets utilized in Pranet~\cite{fan2020pranet}, encompassing five commonly used datasets: Kvasir-SEG, CVC-ClincDB, CVC-ColonDB, EndoScene, and ETIS. 
Specifically, the training set consists of 900 images from Kvasir-SEG and 550 images from ClinicDB. 
The test sets comprise 100 images from Kvasir-SEG, 62 images from CVC-ClincDB, 380 images from CVC-ColonDB, 60 images from EndoScene, and 196 images from ETIS. 
%

\begin{figure*}
  \centering
  \small
  \setlength{\abovecaptionskip}{2pt}
  \begin{overpic}[width=1\textwidth]{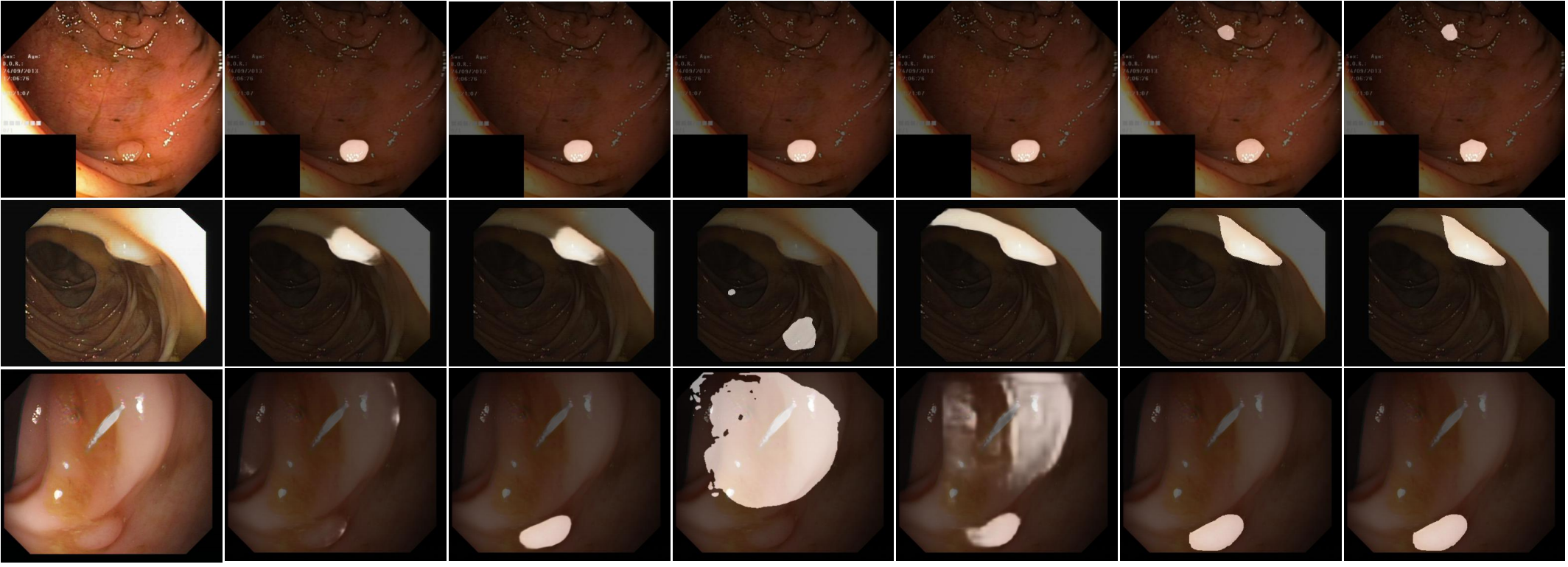}
    \put(5, 37){{\textcolor{black}{Images}}}
    \put(19.5, 37){{\textcolor{black}{U-Net}}}
    \put(33.5, 37){{\textcolor{black}{PraNet}}}
    \put(44.5, 37){{\textcolor{black}{DS-TransUNet}}}
    \put(61, 37){{\textcolor{black}{CaraNet}}}
    \put(76, 37){{\textcolor{black}{\nameofmethod{}}}}
    \put(87.5, 37){{\textcolor{black}{Ground Truth}}}

    \put(-2,27.5){\rotatebox{90}{Kvasir}}
    \put(-2,15 ){\rotatebox{90}{ClinicDB}}
    \put(-2,3.5 ){\rotatebox{90}{ColonDB}}
  \end{overpic}
  \caption{Segmentation results of different methods on Kvasir, ClinicDB and ColonDB dataset.}
  \label{fig:polypresult}
\end{figure*}

\subsection{Implementation Details}

We use PyTorch~\cite{paszke2019pytorch} and mmsegmentation~\cite{contributors2020mmsegmentation}  to implement~\nameofmethod{} for all experiments. 
The input resolution during training is set to $224\times 224$, and the batch size is set to 6.
The number of iterations during training is 80k.
We employ the AdamW optimizer for training with an initial learning rate of 0.0002, a momentum of 0.9, and a weight decay of 1e-4.
All the experiments are conducted on one NVIDIA RTX 3090 GPU. 
Following the configuration of previous research~\cite{lin2022ds}, evaluation metrics mIoU and mDice are employed.
The calculation of flops and parameters in the experiments is based on an input size of $512\times 512$, and the calculation method is from the mmsegmentation project.
%

\subsection{Analysis of Experimental Results}

\myPara{Comparison with the State-of-the-Art Methods.}
We compare the segmentation performance of  \nameofmethod{} with other state-of-the-art models.
Table~\ref{POLYP} presents a comprehensive comparison with CNN-based methods, including U-Net~\cite{ronneberger2015u}, UNet++\cite{zhou2018unet++}, SFA~\cite{fang2019selective}, ACSNet~\cite{zhang2020adaptive}, PraNet~\cite{fan2020pranet} and SANet~\cite{wei2021shallow}.
In addition, we also compare with Transformer-based methods, including TransFuse~\cite{zhang2021transfuse}, HarDNet-MSEG~\cite{huang2021hardnet}, DS-TransUNet~\cite{lin2022ds}, Polyp-PVT~\cite{dong2021polyp}, and SwinE-Net~\cite{park2022swine}, and refinement-based methods, such as CaraNet~\cite{lou2022caranet}, ColonFormer~\cite{thanh2022colonformer}, and SegT~\cite{chen2023segt}.
As depicted in Table~\ref{POLYP}, it is evident that our proposed boundary sensitive method, called~\nameofmethod{}, outperforms other listed methods.
The visualization results are shown in Fig.~\ref{fig:polypresult}, from which it can be observed that our proposed method outperforms the previous methods in boundary processing and the processing of small polyps.

\myPara{Small Polyp Analysis.}
We also evaluate the performance of small polyps.
This type of polyps tends to appear at the onset of the disease and has lower contrast~\cite{antonelli2021impact}.
Specifically, we follow the approach presented in CaraNet~\cite{lou2022caranet} and focus on evaluating polyps that make up less than 6\% of the entire image.
%
In this experiment, we compare with a CNN-based method PraNet~\cite{fan2020pranet}, a Transformer-based method DS-TransUNet~\cite{lin2022ds}, and a refinement-based method CaraNet~\cite{lou2022caranet}.
The results are shown in Fig.~\ref{fig:smallpolyp}.
It can be observed that \nameofmethod{} performs better in small polyps than other methods thanks to the boundary sensitive strategy.
Notably, \nameofmethod{} even outperforms CaraNet, a method specially designed for small polyp targets.

\begin{table}[!tp]\centering
  \small
  \centering
  \setlength{\abovecaptionskip}{2pt}
  \setlength\tabcolsep{3pt}
  \caption{Ablations on different encoders and feature aggregation methods.
  `Polyper w/o BSR' means not
    to refine the initial segmentation results
  }
  \begin{tabular}{lccc}
  \toprule
Encoder  & Decoder     & mIoU   & mDice\\
\midrule
ResNet-50 & Ours w/o BSR  &82.75 &90.13  \\ 
ResNet-50 & Ours w/ BSR   &84.97 \textbf{(2.22 $\uparrow$)} &91.29 \textbf{(1.16 $\uparrow$)} \\
\midrule
MiT-B1    & Ours w/o BSR  &87.74 &93.28 \\ 
MiT-B1    & Ours w/ BSR   &89.19 \textbf{(1.45 $\uparrow$)} &94.13 \textbf{(0.85 $\uparrow$)}	 \\ 
\midrule
Swin-T  &Non-Local w/o BSR  &84.56  &91.31 \\
Swin-T  &Non-Local w/ BSR   &86.55 \textbf{(1.99 $\uparrow$)}  &92.55 \textbf{(1.24 $\uparrow$)} \\
\midrule
Swin-T  &Hamburger w/o BSR  &82.40  &89.85 \\ 
Swin-T  &Hamburger w/ BSR   &83.67 \textbf{(1.27 $\uparrow$)}  &90.65 \textbf{(0.80 $\uparrow$)} \\ 
\midrule
Swin-T       & Ours w/o BSR  &87.12 &92.90 \\
Swin-T       & Ours w/ BSR  &90.57 \textbf{(3.45$\uparrow$)}  &94.49 \textbf{(1.59$\uparrow$)} \\ 
\bottomrule
\end{tabular}
  \label{Ablation_endocer}
\end{table}

\subsection{Ablation Study}
We conducte extensive ablation experiments on the Kvasir dataset to analyze~\nameofmethod{}.

\myPara{Ablations on Encoder.}
Initially, we conduct experiments to evaluate the impact of different encoders.
We choose the commonly used ResNet-50~\cite{he2016deep} and MiT-B1~\cite{xie2021segformer} as the evaluation encoders. 
As shown in Table~\ref{Ablation_endocer}, when using our decoder and ResNet-50 as an encoder, a significant enhancement of 2.22 for mIoU and 1.16 for mDice is observed compared to not refining the initial segmentation results.
%
Additionally, using the MiT-B1 as encoder brings significant enhancement.
This demonstrates the broad applicability of~\nameofmethod{} to various encoders.

\begin{table}[tp]
  \centering
  \small
  \setlength{\abovecaptionskip}{2pt}
    \setlength\tabcolsep{7.5pt}
 \caption{Ablations on~\nameofedge{}.
  Here, `Number of iterations' represents the number of iterations applied by the erosion operator.}
  \begin{tabular}{lcccccccc}
  \toprule
  \multirow{2}{*}{RS}  &\multicolumn{6}{l}{\makecell{Number of iterations}} &\multirow{2}{*}{mIoU} &\multirow{2}{*}{mDice}   \\ \cmidrule(lr){2-7}
   & \makecell{1}  & \makecell{2} & \makecell{3} & \makecell{4} & \makecell{5} & \makecell{6}     \\ \midrule
  \multicolumn{1}{c}{}  & & & & & & &86.55 &92.57\\
 \multicolumn{1}{c}{\cmark}  &\cmark & & & & & &87.14 &92.91\\
  \multicolumn{1}{c}{\cmark}   & &\cmark & & & & &87.95 &93.40\\
   \multicolumn{1}{c}{\cmark}   & & &\cmark & & & &88.79 &93.90\\
   \multicolumn{1}{c}{\cmark}   & & & & &\cmark & &89.66 &94.42\\
   \multicolumn{1}{c}{\cmark}   & & & & & &\cmark &88.79 &93.30\\
\midrule
 \multicolumn{1}{c}{\cmark}   & & & &\cmark & &  &\textbf{90.57} &\textbf{94.49}\\
\bottomrule 
\end{tabular}
\label{Ablation_EE}
\end{table}

\begin{figure*}
  \centering
  \small
  \setlength{\abovecaptionskip}{2pt}
  \begin{overpic}[width=1\textwidth]{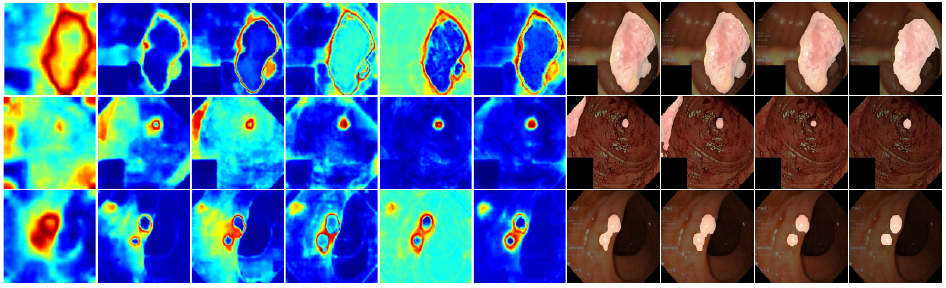}
        \put(4, 31.5){{\textcolor{black}{(a)}}}
        \put(14, 31.5){{\textcolor{black}{(b)}}}
        \put(24, 31.5){{\textcolor{black}{(c)}}}
        \put(34, 31.5){{\textcolor{black}{(d)}}}
        \put(44, 31.5){{\textcolor{black}{(e)}}}
        \put(54, 31.5){{\textcolor{black}{(f)}}}
        \put(64, 31.5){{\textcolor{black}{(g)}}}
        \put(74, 31.5){{\textcolor{black}{(h)}}}
        \put(84, 31.5){{\textcolor{black}{(i)}}}
        \put(94, 31.5){{\textcolor{black}{(j)}}}
  \end{overpic}
  \caption{Visual analysis of the feature maps produced by different versions of \nameofmethod{}. All images are selected from the Kvasir dataset. (a) w/o Boundary Sensitive Refinement; (b) w/o Region Separation; (c) w/o Boundary Sensitive Attention;(d) \nameofmethod{} w/o channel attention; (e) w/o spatial attention; (f) \nameofmethod{}; (g) segmentation results w/o Boundary Sensitive ; (h) segmentation results  w/o  Region Separation ; (i) segmentation results w/ \nameofmethod{}; (j) ground-truth annotations.}
  \label{fig:attentionmap}
\end{figure*}
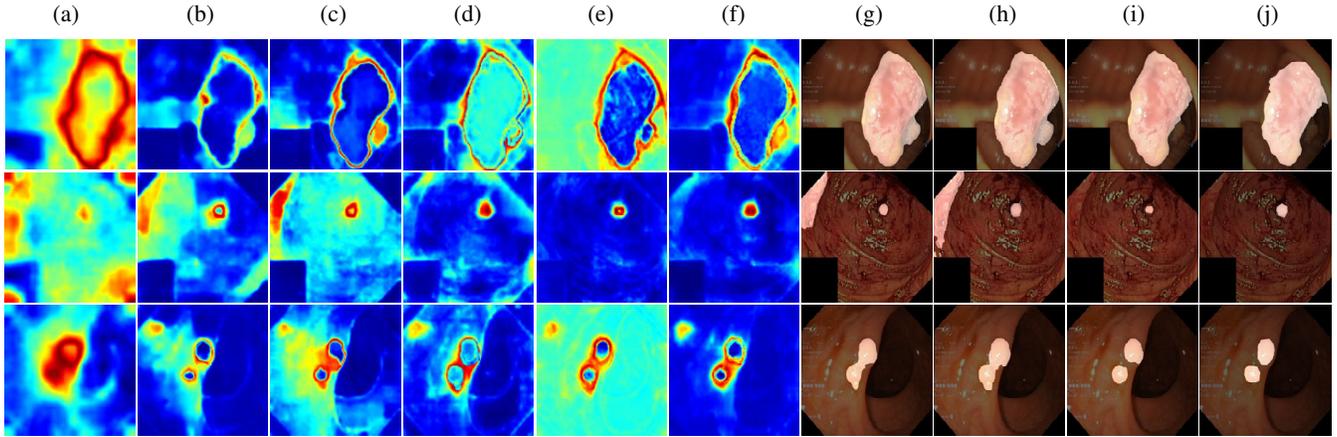

\myPara{Ablations on Potential Boundary Extraction.} We first evaluate the influence of different feature aggregation methods. 
For this evaluation, we consider the Non-Local block~\cite{wang2018non} and Hamburger~\cite{geng2021attention}. 
%
As Table~\ref{Ablation_endocer} shows, our proposed boundary sensitive approach~\nameofmethod{} is compatible with different feature aggregation methods, reflecting our method's generalization ability.
Furthermore, it is noticeable that the Non-Local block and Hamburger do not surpass the performance of our proposed feature aggregation method when used for feature aggregation.
We attribute this to the fact that these two methods are originally designed for semantic segmentation in natural images.
%
Given the limited amount of medical data available for segmentation and the resulting challenges of network convergence, their performance in medical segmentation remains unsatisfactory.

Furthermore, we conduct experiments to analyze the effectiveness of Region Separation (RS). 
%
%
In the absence of RS, the subsequent refinement stage employs the full mask to encompass the entire foreground area.
As presented in Table~\ref{Ablation_EE}, refinement with RS demonstrates improvements of 4.40 on mIoU and 2.59 on mDice compared to refinement with the whole mask. 
%
This is because the boundary regions of the initial segmentation results contain unreliable features with low confidence.
When the whole foreground is refined, it is affected by these unreliable features, which reduces the quality of the segmentation results.

Finally, we conduct experiments on the width of the boundary region, and the results are shown in Table~\ref{Ablation_EE}. 
The width of the boundary region is determined by performing a subtraction operation between the mask after applying the erosion operator and the mask after applying the dilation operator, following~\cite{zhu2023optimal} to calculate.
%
%
%
For this experiment, our evaluation focuses on assessing the impact of the number of iterations executed by the erosion operator.
From the table, it can be observed that the optimal number of iterations for the erosion operator is 4.
We suggest that the readers use this number in their experiments.

\myPara{Ablations on Boundary Sensitive Refinement.}
Here, we conduct experiments to validate the importance of boundary sensitive refinement stage.
First, we evaluate the importance of the two branches, spatial attention and channel attention.
Table~\ref{Ablation_CAR} clearly illustrates the contributions of spatial attention and channel attention.
This illustrates the effectiveness of fully utilizing the relationship between the boundary region and the interior polyp region and the relationship between the interior polyp region and the background region for producing polyp regions with accurate boundaries.
%

\begin{table}[!tp]
    \centering
    \small
    \setlength{\abovecaptionskip}{2pt}
    \centering
    \setlength\tabcolsep{14pt}
    \caption{Ablations on the boundary sensitive attention module.
    `SA' means spatial attention.
    `CA' means channel attention.}
    \label{Ablation_CAR}
    \begin{tabular}{lcccc} \toprule
  SA  &CA & mIoU & mDice\\
 \midrule
     &    &87.65  &93.22   \\
  \cmark  & &89.45 \textbf{(1.80 $\uparrow$)}  &94.28 \textbf{(1.06 $\uparrow$)}   \\
    &\cmark  &89.16 \textbf{(1.51 $\uparrow$)}  &94.12 \textbf{(0.90 $\uparrow$)}   \\
\cmark  &\cmark  &90.57 \textbf{(2.92$\uparrow$)}   &94.49 \textbf{(1.27$\uparrow$)}	  \\ 
    \bottomrule
    \end{tabular}
\end{table}

\begin{table}[!tp]
    \centering
    \small
    \setlength{\abovecaptionskip}{2pt}
    \centering
    \setlength\tabcolsep{7.5pt}
    \caption{Ablations on full-stage sensitive strategy. When all features are used, the performance is the best.}
    \begin{tabular}{lcccccccc}  \toprule
    ${D_3}$  & ${D_2}$ & ${D_1}$  & ${D_0}$ & mIoU & mDice \\
    \midrule
    \cmark   &       &         &       &85.43  &91.88   \\ 
    \cmark   &\cmark &         &       &86.26 \textbf{(0.83 $\uparrow$)}  &92.16 \textbf{(0.28 $\uparrow$)}   \\ 
    \cmark   &\cmark &\cmark   &       &88.90 \textbf{(2.64 $\uparrow$)}  &93.48 \textbf{(1.32 $\uparrow$)}  \\ 
    \cmark   &\cmark &\cmark   &\cmark &90.57 \textbf{(1.67 $\uparrow$)}  &94.49 \textbf{(1.01 $\uparrow$)}   \\ 
    \bottomrule
    \end{tabular}
    \label{Ablation_FSS}
\end{table}

Then, we evaluate the importance of our full-stage sensitive strategy. 
The purpose of this experiment is to verify whether it is necessary to refine all the features in the feature aggregation part.
Table~\ref{Ablation_FSS} illustrates the effectiveness of fully utilizing different levels of features to improve the quality of segmentation results with accurate boundaries.
We can see that gradually incorporating more features from lower levels can continuously increase the model's performance.

\myPara{Visual Analysis.}
To gain a more comprehensive understanding of the functionality of each component in our proposed \nameofmethod{}, we utilize the visualization method proposed in~\cite{komodakis2017paying} to visualize the output feature maps generated by different versions of \nameofmethod{}.
The visual results are presented in Fig.~\ref{fig:attentionmap}.
From Fig. ~\ref{fig:attentionmap}(b), it can be observed that when refining the initial results from a global perspective, this approach has limited effect and does not solve the problem of edge blurring due to the presence of interference in the boundary regions with low-confidence predictions.
In contrast, as can be observed from Fig.~\ref{fig:attentionmap}(f), our method is effective when augmented by modeling the relationship between the interior polyp region and the boundary region and the relationship between the interior polyp region and the background region to differentiate the boundaries of the lesion regions.

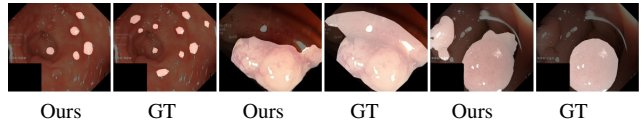
\begin{figure}
    \centering
    \small
  \begin{overpic}[width=1\linewidth]{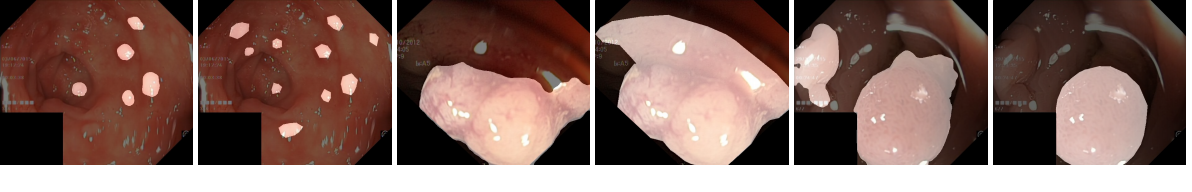}
\put(5, -2){{\fontsize{8}{10}\selectfont \textcolor{black}{Ours}}}
\put(22, -2){{\fontsize{8}{10}\selectfont \textcolor{black}{GT}}}
\put(37, -2){{\fontsize{8}{10}\selectfont \textcolor{black}{Ours}}}
\put(54, -2){{\fontsize{8}{10}\selectfont \textcolor{black}{GT}}}
\put(70, -2){{\fontsize{8}{10}\selectfont \textcolor{black}{Ours}}}
\put(87, -2){{\fontsize{8}{10}\selectfont \textcolor{black}{GT}}}
  \end{overpic}
	\caption{Failure cases of~\nameofmethod{}.
 }
	\label{fig:failure}
\end{figure}

\myPara{Limitations of~\nameofmethod{}.}
We show some failure cases of~\nameofmethod{} in Fig.~\ref{fig:failure}. 
First, we assume polyp localization is accurate and cannot handle false positives or
false negatives well.
Second, we employ the fixed-width method to define the boundary width and extract boundary regions.
This may not account for the diversity of polyp features.
In the future, we will explore adaptive boundary width methods.

\section{Conclusion}
We present \nameofmethod{}, a novel approach for polyp segmentation.
%
We employ morphology operators to delineate boundary and interior polyp regions from the initial segmentation results.
Then, we leverage the features of the interior polyp regions to enhance the features of boundary regions.
Our experiments on five datasets demonstrate the remarkable performance of \nameofmethod{}.

\myPara{Acknowledgements}
This research was supported by NSFC (No. 62276145), the Fundamental Research Funds for the Central Universities (Nankai University, 070-63223049), CAST through Young Elite Scientist Sponsorship
Program (No. YESS20210377). Computations were supported by the Supercomputing Center of Nankai University (NKSC).

\bibliography{aaai24}

\end{document}